\def\year{2021}\relax
\documentclass[letterpaper]{article} 
\usepackage{aaai21}  
\usepackage{times}  
\usepackage{helvet} 
\usepackage{courier}  
\usepackage[hyphens]{url}  
\usepackage{graphicx} 
\usepackage{makecell}
\usepackage{natbib}  
\usepackage{caption} 
\title{Medication Error Detection Using Contextual Language Models}
\author {
    Yu Jiang\textsuperscript{\rm 1} and
    Christian Poellabauer\textsuperscript{\rm 2} \\
}
\affiliations {
    \textsuperscript{\rm 1} University of Notre Dame, Notre Dame, IN, USA \quad
    \textsuperscript{\rm 2} Florida International University, Miami, FL, USA \\
    yjiang7@nd.edu, cpoellab@fiu.edu
}

\begin{document}

\maketitle

\begin{abstract}
Medication errors most commonly occur at the ordering or prescribing stage, potentially leading to medical complications and poor health outcomes. While it is possible to catch these errors using different techniques; the focus of this work is on textual and contextual analysis of prescription information to detect and prevent potential medication errors. In this paper, we demonstrate how to use BERT-based contextual language models to detect anomalies in written or spoken text based on a data set extracted from real-world medical data of thousands of patient records. The proposed models are able to learn patterns of text dependency and predict erroneous output based on contextual information such as patient data. The experimental results yield accuracy up to 96.63\% for text input and up to 79.55\% for speech input, which is satisfactory for most real-world applications.

\end{abstract}

\section{Introduction}
%

Over 6,800 prescription medication types are available in the U.S. alone; each year hundreds of thousands of patients experience adverse reactions or complications and between 7,000 and 9,000 people die each year due to medication error~\cite{tariq2021}. The leading type of error is dosing error, followed by omissions and wrong drug types~\cite{mulac2021} and most of these errors occur at the ordering or prescribing stages. Once a prescription has been recorded, e.g., in the electronic health records (EHR) of a patient, textual analysis can be used to detect such errors and prevent their consequences. Physicians are also increasingly rely on automatic speech recognition (ASR) systems to communicate with medical or computing equipment, including to order drugs~\cite{Latif2021}. Given that ASRs cannot provide perfect transcriptions at all times, they become another source of prescription errors. These erroneous prescriptions could have severe consequences on patient treatment~\cite{oura2021}, e.g., by administering the wrong type or amount of drug~\cite{rodziewicz2020}. 




\begin{table}[t]
\centering
\begin{tabular}{|c|c|c|}
\hline
\textbf{Type} & \textbf{Content} & \textbf{Valid?} \\
\hline
EHR & \makecell[l]{\textbf{Diagnosis}: Cholecystitis \\ \textbf{Prescription}: Aspirin eighty one \\ milligrams daily} & Yes \\
\hline
EHR & \makecell[l]{\textbf{Diagnosis}: Bacteremia endocarditis \\ \textbf{Prescription}: Lisinopril five millig- \\ rams oral administration once a day} & No \\
\hline
ASR & \makecell[l]{\textbf{Diagnosis}: Peripheral vascular \\ disease \\ \textbf{Prescription}: Blood pressure \\ (lopressor) seventy five milligrams \\ orally administration} & No \\
\hline
\end{tabular}
\caption{Examples of prescriptions from EHR and ASR.}
\label{tab:exp}
\end{table}

\begin{figure*}[t]
\centering
\includegraphics[scale=0.3]{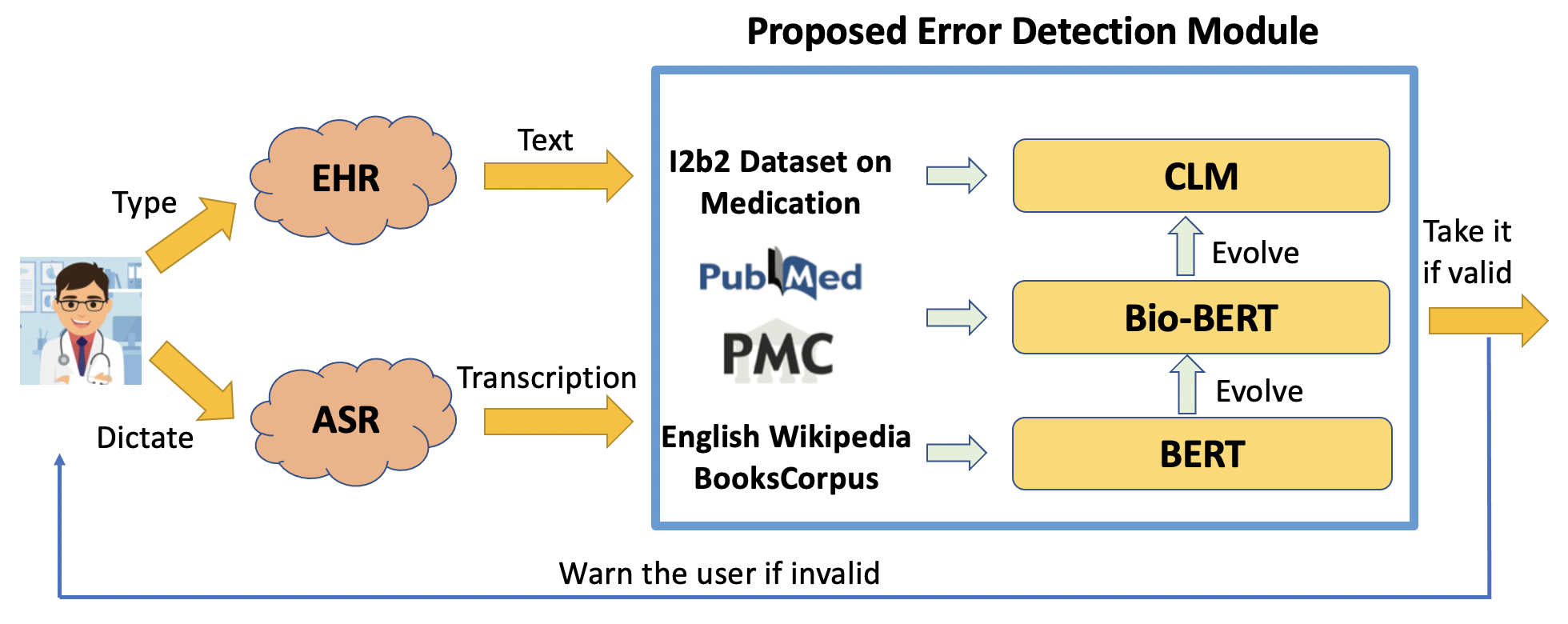} 
\caption{Detecting prescription errors using a contextual language model: the error detection module receives text or speech input from the physician and decides the validity of the prescription. The proposed CLM model is the key to error detection, which is based on BERT and Bio-BERT, and trained with in-domain corpora.}
\label{fig:sys}
\end{figure*}

In this paper, we propose to detect invalid prescriptions by verifying that it is an appropriate choice given the status (or context) of the patient receiving the medication. Toward this end, we already have a lot of relevant information available, e.g., using a hospital's digital systems that contain diagnoses, health history, medication history, etc., of patients. In this paper, we leverage such {\em contextual knowledge} to detect potential prescription (from textual information provided by a physician) or transcription errors (from an ASR). This is demonstrated by the examples shown in Table~\ref{tab:exp}. Human error could lead to an invalid prescription which will be stored in an EHR together with other patient-specific data. The first row of the table shows a valid prescription, i.e., the drug type, usage, etc., is typical for the patient's diagnosis. In the second row of the table, Lisinopril is not an appropriate drug for Bacteremia Endocarditis and this shoulg be flagged as invalid. Finally, the third row shows the transcription output of an ASR where the drug name is mis-interpreted by the system (``blood pressure'' instead of ''Lopressor''), making this entry again invalid. In order to detect erroneous entries, we propose to use a {\bf contextual language model (CLM)} that utilizes context information about a patient to determine whether a prescription is valid or not. The basic idea is to analyze the correlation between prescription and the context. If they are highly correlated, our system will accept it. On the other hand, the system will trigger an alert to the user, allowing the user to review and correct the prescription. In this work, the proposed CLM is a neural network based on two pre-trained word representations: BERT~\cite{Devlin2019BERTPO} and BioBERT~\cite{lee2020biobert}, which are described in the remainder of this paper. 

\section{Proposed Methodology}

The goal of our work is to prevent medical errors from happening with the help of contextual information, i.e., detecting incorrect prescriptions using the proposed contextual language model. We can consider two types of sources for the prescription. First, a medical professional types a prescription into a system and any errors at this point are true human errors, e.g., the physician misjudging the patient's situation or not having access to all required information (e.g., patient's health history or allergies, etc.). Second, a physician dictates the drug information and an ASR system translates the recording into text. In the latter case, errors can also be due to transcription errors by the ASR. These two scenarios are demonstrated in Figure~\ref{fig:sys}.

\subsection{Problem Formalization}
We define the set of medical inputs as $H$, and the set of contextual knowledge as $C$. Each element $h_i$ in $H$ has its corresponding contextual knowledge $c_i$, where $h_i \in H$ and $c_i \in C$. Both $h_i$ and $c_i$ are text, consisting of a sequence of words. The problem we intend to solve here is to determine whether $h_i$ is correct or not based on its corresponding context $c_i$. The proposed method is to calculate the likelihood of $h_i$ under the condition of $c_i$ $P(h_i|c_i)$, and then make a decision according to that value. The likelihood is calculated by the CLM, where the CLM is trained using labeled pairs $(h_i, c_i)$. A label represents the correlation between $h_i$ and $c_i$; if they are correlated, the label would be positive, and negative if uncorrelated. Therefore, the trained CLM could output the likelihood $P(h_i|c_i)$, measuring the correlation between $h_i$ and $c_i$, which approaches ``1'' when correlation is high and ``0'' when correlation is low.

When an ASR system is used, the spoken form of $h_i$ is the input to the ASR system, which produces the transcription $h^{'}_{i}$. Then, the pair of text $(h^{'}_{i}, c_i)$ is the input to the CLM. Note that $h^{'}_{i}$ could be erroneous compared to $h_i$. In this situation, the performance of CLM will decrease, because CLM is trained using $(h_i, c_i)$, and $h^{'}_{i}$ is likely to follow different patterns from $h_i$.

\begin{figure}[t]
\centering
\includegraphics[width=0.8\columnwidth]{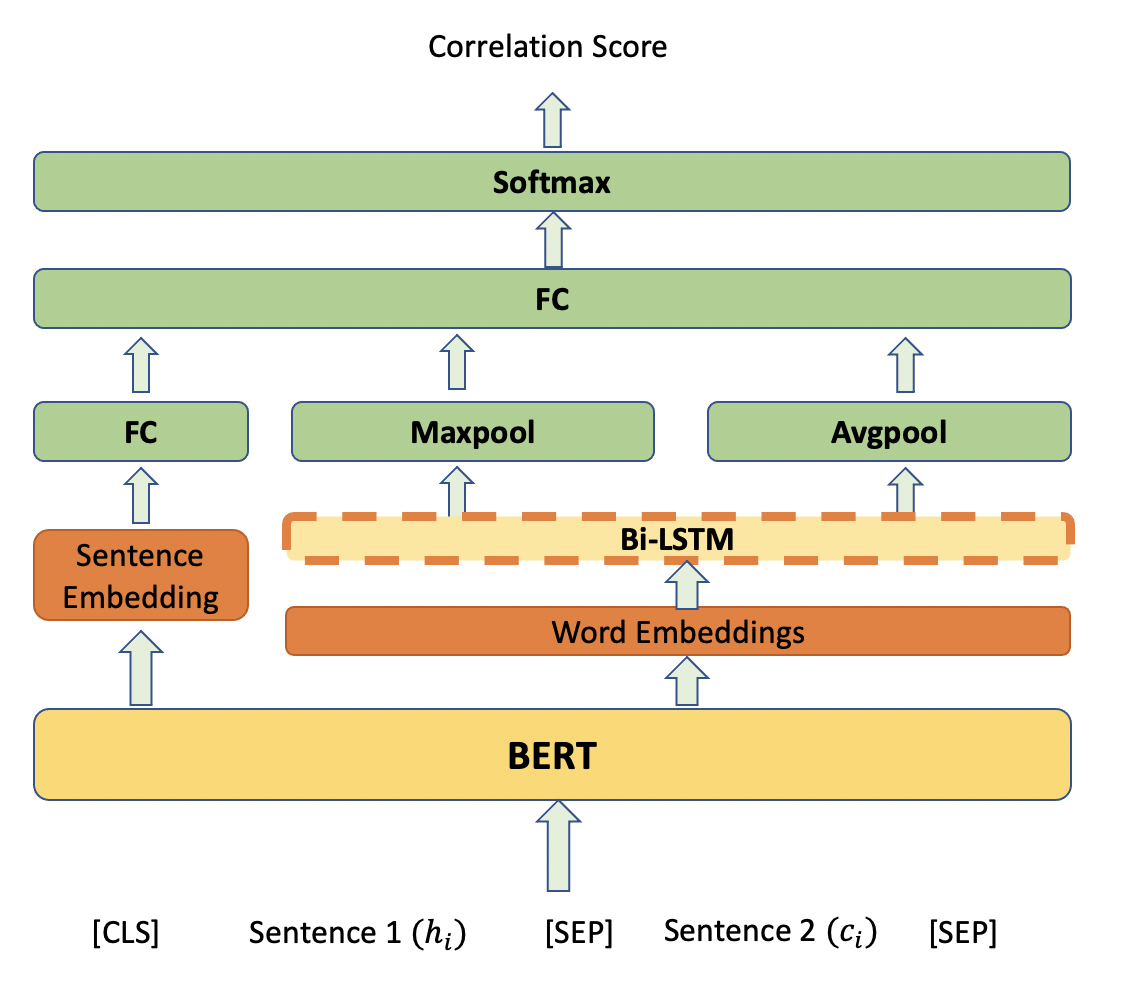} 
\caption{BERT-based contextual language model: Bi-LSTM can be added depending on the configuration.}
\label{fig:m1}
\end{figure}

\subsection{Contextual Language Models}

The CLM takes text as input, i.e., we need to convert the sequence of words into vectors. There are several methods of learning word representations from a large amount of un-annotated text, such as Word2vec~\cite{mikolov2013}, ELMo~\cite{peters2018}, and CoVe~\cite{mccann2017}. In the proposed approach, we build the CLM using BERT~\cite{Devlin2019BERTPO} and BioBERT~\cite{lee2020biobert}, because BERT achieves state-of-the-art performance for most NLP tasks and BioBERT is a model (based on BERT) specifically for the biomedical field. They share the same structure of bidirectional Transformer encoder~\cite{vaswani2017}, but are trained using different corpora. The model architecture is a stack of 6 identical layers and each layer has two sub-layers. One is a multi-head self-attention mechanism, the other is a simple fully connected feed-forward network. There are also residual connection and layer normalization steps after each sub-layer. The BERT model is pre-trained using two unsupervised tasks, i.e., Masked Language Model (MLM) and Next Sentence Prediction (NSP), using English Wikipedia of 2.5 billion words and BooksCorpus of 0.8 billion words. BioBERT is further trained on biomedical text, i.e., PubMed abstracts of 4.5 billion words and PMC full-text articles of 13.5 billion words, both of which contain a large number of domain-specific terms. The BERT and BioBERT models are initialized using the weights after pre-training. 

The input format to the CLM is consistent with BERT and BioBERT. The WordPiece embedding~\cite{wu2016googles} is used to deal with the out-of-vocabulary problem. Two special symbols are added to each word sequence, i.e., the CLS and SEP token. The CLS is added to the head, and the final corresponding hidden state is the whole sequence representation. The SEP token is added to the end of each sentence, in order to separate the sentences. We directly fine-tune the BERT model as the baseline. A linear head is used to map the sentence representation to classification label. The shortcoming here is that it does not take advantage of all the information output from BERT. We could otherwise make use of all the output word representations. Therefore, we propose the model shown in Figure~\ref{fig:m1}. Besides using sentence embedding, we also take the word embedding after max-pooling and average-pooling as feature outputs from BERT. Furthermore, a Bi-LSTM layer could be added in order to further extract features from the word embedding. 

\section{Experimentation}

\begin{figure}[t]
\centering
\includegraphics[width=0.9\columnwidth]{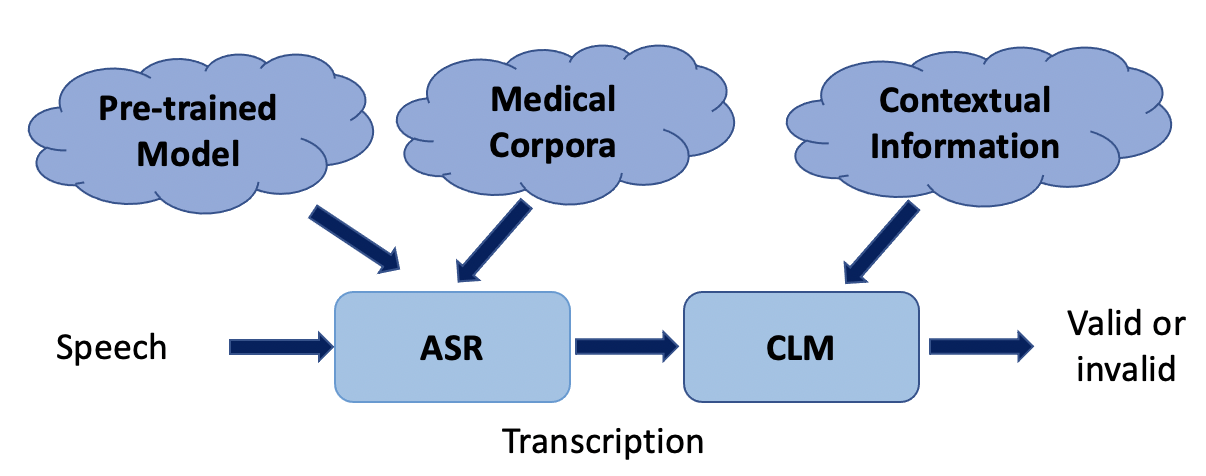} 
\caption{Experimental framework for speech input.}
\label{fig1}
\end{figure}

We perform experiments using two types of data: speech and text. When the data type is text, the medical prescription would be directly input to the CLM. And when the input is speech, the experimental framework is shown in Figure~\ref{fig1}. First we input the utterance to the ASR (which is constructed using a pre-trained model for normal language such as ASPIRE, and medical corpora, which provide domain-specific information, such as the medical dictionary and the medical language model), and obtain the decoding results. Then the CLM will determine whether the medical input is correct or not based on the contextual information.

\begin{table}[t]
\centering
\setlength\tabcolsep{3.5pt}
\begin{tabular}{|l|c|c|c|c|}
\hline
\textbf{Model} & \textbf{Accuracy} & \textbf{Precision} & \textbf{Recall} & \textbf{F1} \\
\hline
$BERT$ & 0.9625 & 0.9354 & 0.9928 &  0.9633 \\
\hline
$BERT_{mlp}$ & 0.9647 & 0.9404 & 0.9916 & 0.9653 \\
\hline
$CLM$ & \textbf{0.9663} & \textbf{0.9425} & \textbf{0.9924} & \textbf{0.9668} \\
\hline
$CLM_{lstm}$ & 0.9645 & 0.9395 & 0.9922 & 0.9651 \\
\hline
$CLM_{bio}$ & 0.9633 & 0.9374 & 0.9921 & 0.9640 \\
\hline
$CLM_{biolstm}$ & 0.9653 & 0.9397 & 0.9935 & 0.9659 \\
\hline 

\end{tabular}
\caption{Performance of different models for text input.}
\label{tab1}
\end{table}

\begin{table}[t]
\centering
\setlength\tabcolsep{3.5pt}
\begin{tabular}{|l|c|c|c|c|}
\hline
\textbf{Model} & \textbf{Accuracy} & \textbf{Precision} & \textbf{Recall} & \textbf{F1} \\
\hline
$BERT$ & 0.7659 & 0.7749 & 0.5004 & 0.6081 \\
\hline
$BERT_{mlp}$ & 0.7724 & 0.7743 & 0.5267 & 0.6269 \\
\hline
$CLM$ & 0.7759 & 0.7759 & 0.5380 & 0.6355 \\
\hline
$CLM_{lstm}$ & 0.7579 & \textbf{0.7937} & 0.4503 & 0.5746 \\
\hline
$CLM_{bio}$ & \textbf{0.7955} & 0.7879 & \textbf{0.5976} & \textbf{0.6797} \\
\hline
$CLM_{biolstm}$ & 0.7508 & 0.7645 & 0.4530 & 0.5689 \\
\hline 

\end{tabular}
\caption{Performance of different models for speech input.}
\label{tab2}
\end{table}

\subsection{Dataset Generation}
We use the dataset of the 2019 National NLP Clinical Challenges (n2c2) on Medication, provided by the i2b2 Center. The original set includes about 2,000 patients' hospitalization records, which provides information such as diagnosis, illness history, laboratory data, hospital course, and discharge medications. We extract the contents of diagnosis and discharge medications in each record, and divide the whole paragraph of medications into different items. Each item of medication has its corresponding diagnosis, and we believe they are highly correlated. The number of this kind of correlated pairs is 8,621. Then we further clean the pairs by removing serial numbers and punctuation, deleting out-of-domain items, handling special symbols, and replacing abbreviations with full names. The final number of pairs is then 6,901. 

For a binary classification task, we not only need correlated pairs, but also uncorrelated pairs. Therefore, we create a dataset of uncorrelated pairs by randomly combining diagnoses and medications into different pairs. Many medications can be used for different disorders; therefore, we design an algorithm to calculate a distance between diagnoses. If the distance is greater than a threshold, we just discard this pair. The number of generated uncorrelated pairs is 70,000. To balance the data with correlated and uncorrelated items, we further duplicate the correlated pairs 10 times to obtain 69,010 pairs.

The data for CLM is prepared by splitting the entire set into training, validation, and testing sets in the proportion of 6:2:2. The correlated pairs have positive labels and uncorrelated pairs have negative labels. 

The dataset only includes text data; the speech input is processed using Google's Text-to-Speech (TTS) API. The medical inputs $H$ in the test set are converted from speech to text and then used as input to the ASR system for decoding. 

\subsection{ASR Implementation}
The ASR system is implemented using the Kaldi Speech Recognition Toolkit~\cite{Povey2011}. It is hard to train a good ASR model from scratch, because large amounts of speech and text data for the specific domain will be required. Instead, we adapt the ASPIRE model to the medical domain by simply combining the ASPIRE language model and the medical language model trained from our i2b2 dataset. The decoding utility in Kaldi used in this work is ``online2- wav-nnet3-latgen-faster'', which provides fast and accurate decoding.


\subsection{Results and Analysis}

We conduct experiments on the following models: BERT, BERT with a multilayer perceptron (MLP) head ($BERT_{mlp}$), BERT-based CLM ($CLM$), BERT-based CLM with Bi-LSTM ($CLM_{lstm}$), BioBERT-based CLM ($CLM_{bio}$), and BioBERT-based CLM with Bi-LSTM ($CLM_{biolstm}$). All models are initialized the same way and are trained to their best states. For a binary classification task, the metrics are accuracy, precision, recall, and F1 score. We take 0.5 as the threshold. For text input, the results on the test set are shown in Table~\ref{tab1}. These results show that the BERT-cased CLM achieves the best performance with 96.63\% accuracy of the test set, and 0.4\% absolute increment compared to the baseline. We have a large test set of nearly 30 thousand, so about a thousand wrong predictions made by BERT become correct using our best model.

We further explore the performance of different models for speech input. The input to the CLM becomes the hypotheses output from the ASR. In this case, we need to note that the ASR sometimes makes mistakes, and in our experiments, the word error rate (WER) is 28.69\%. Therefore, the medical commands, which at first are correlated with the context, might become uncorrelated after ASR decoding, and the labels of test set have to be modified. We determine the labels through comparing the text before and after ASR. If some essential entities, such as the medication name, dosage, or usage are incorrect, we change the label from ``1'' to ``0''. In the test set, we have 13,755 correlated pairs at the beginning, and 10,092 correlated pairs left after processing. The threshold is still 0.5. The results are shown in Figure~\ref{tab2}. From the results we see that when there is an ASR system involved, the BioBERT-based CLM performs the best with an accuracy of 79.55\% and a 2.96\% absolute increment compared to the baseline. The erroneous transcriptions by ASR negatively impact the models, but applying BioBERT can minimize these impacts. Moreover, adding the Bi-LSTM layer and making the network deeper does not improve the performance.

\section{Conclusions}

This paper proposes a solution to detecting prescription errors with the help of contextual language models (CLM). This kind of models are effective in dealing with text data, and are able to make validity predictions based on the correlations. We achieve the best accuracy of 96.63\% using the BERT based CLM when the physicians type prescription data directly into the EHR, and 79.55\% using the BioBERT based CLM when the physicians dictate the prescriptions using an ASR. Future work might include integrating more diverse contextual knowledge, i.e., in addition to patient data, we can also consider physician preferences, databases that describe adverse reactions, clinical workflows and recommender outputs, etc.

\bibliography{ref.bib}
\end{document}